# Enabling Building Information Model-Driven Human-Robot Collaborative Construction Workflows with Closed-Loop Digital Twins

Xi Wang[1*], Hongrui Yu[2*], Wes McGee[3], Carol C. Menassa[4†], Vineet R. Kamat[5]

[1]Assistant Professor, Department of Construction Science, Texas A&M University, 3337 TAMU, College Station, TX 77843, USA

[2]Ph.D. Candidate, Department of Civil and Environmental Engineering, University of Michigan, 2350 Hayward Street, Ann Arbor, MI 48109, USA

[3]Associate Professor, Taubman College of Architecture and Urban Planning, University of Michigan, 2000 Bonisteel Boulevard, Ann Arbor, MI 48109, USA

[4]Professor, Department of Civil and Environmental Engineering, University of Michigan, 2350 Hayward Street, Ann Arbor, MI 48109, USA

[5]Professor, Department of Civil and Environmental Engineering, University of Michigan, 2350 Hayward Street, Ann Arbor, MI 48109, USA

---

[*] These two authors contributed equally to this work.
[†] Corresponding author (menassa@umich.edu).

# Enabling Building Information Model-Driven Human-Robot Collaborative Construction Workflows with Closed-Loop Digital Twins

**ABSTRACT**

The introduction of assistive construction robots can significantly alleviate physical demands on construction workers while enhancing both the productivity and safety of construction projects. Leveraging a Building Information Model (BIM) offers a natural and promising approach to driving robotic construction workflows. However, because of uncertainties inherent in construction sites, such as discrepancies between the as-designed and as-built components, robots cannot solely rely on a BIM to plan and perform field construction work. Human workers are adept at improvising alternative plans with their creativity and experience and thus can assist robots in overcoming uncertainties and performing construction work successfully. In such scenarios, it is critical to continuously update the BIM as work processes unfold so that it includes as-built information for the ensuing construction and maintenance tasks. This research introduces an interactive closed-loop digital twin framework that integrates a BIM into human-robot collaborative construction workflows. The robot's functions are primarily driven by the BIM, but it adaptively adjusts its plans based on actual site conditions, while the human co-worker oversees and supervises the process. When necessary, the human co-worker intervenes in the robot's plan by changing the task sequence or workspace geometry or requesting a new motion plan to help the robot overcome the encountered uncertainties. A drywall installation case study is conducted to verify the proposed workflow. In addition, experiments are carried out to evaluate the system performance using an industrial robotic arm in a research laboratory setting that mimics a construction site and in the Gazebo simulation. Integrating the flexibility of human workers and the autonomy and accuracy afforded by the BIM, the proposed framework offers significant promise of increasing the robustness of construction robots in the performance of field construction work.

**KEYWORDS**

Construction robotics, Building Information Model, Digital twin, Human-robot collaboration.

## 1 INTRODUCTION

Construction has been ill-famed for its dangerous and harsh working environments and excessive physical demands on workers, which often result in a lack of motivation for people, especially those of diverse abilities, to pursue their careers in the industry [1–4]. As a result, the construction industry is facing severe shortages of skilled labor [5,6]. According to the Associated General Contractors of America, 73% of contractors consider worker shortage as their biggest concern in 2022 [7].

Industrial robotic manipulators can exert high physical power and operate at high speeds, and thus have significant potential to reduce the physical burden on human workers [8]. Robots have already been adopted in several industry sectors such as manufacturing, nuclear, healthcare, and rescue to reduce human workers' workload and their exposure to potential hazards [9–13]. Construction automation powered by robotics has demonstrated the potential to reduce the physical demand on construction workers and improve the diversity and inclusion of the construction workforce, thus mitigating the labor shortage issues faced by the industry [14–16]. On the other hand, while robotics is still in its emerging phase in the field of construction automation, the use of Building Information Models (BIMs) has been widely adopted in the industry to support digital construction project workflows [17,18].

A BIM is "a digital representation of physical and functional characteristics of a facility" [19]. It contains a variety of geometric and attribute information, such as 3D models, schedules, construction methods, and materials, which are used to facilitate the construction processes [20]. Although BIMs play important roles in design, communication, and project management throughout the project life cycle [17,21], they generally lack the interoperability needed to support construction robot task planning [22,23]. Currently, robot installation sequences and poses are generated primarily by retrieving geometric data from BIMs encoded using Industry Foundation Classes (IFC) [20,23,24]. However, such proposed approaches are limited to specific construction tasks or types of components, such as bricklaying or wall panels. Moreover, working environments for field construction involve a lot of uncertainties (e.g., deviations in as-built components) that can cause robot failure when following a rigid program. While a BIM can provide information to the robot, the workflow needs the ability to improvise (i.e., dynamically adjust plans based on encountered situations) to flexibly perform the work during the field construction process [25].

Compared to robots, humans are more adept at creative and adaptive planning based on their experience [26,27]. They can adjust a task plan according to what they observe on the construction site to ensure the quality and continuity of the work. Thus, human expertise in improvisation is indispensable for field construction that involves considerable uncertainties and is necessary to support robotic construction [28,29]. In addition, human workers can supervise the robotic construction process to ensure collision-free safe manipulation in dynamic on-site working environments. Therefore, by enabling Human-Robot Collaboration (HRC), the flexibility and robustness of BIM-driven robotic construction systems can be significantly improved.

This research proposes a closed-loop digital twin framework to enable BIM-driven Human-Robot Collaborative Construction (HRCC) workflows, as shown in Figure 1. The system is built upon an interactive and immersive process-level digital twin (I2PL-DT) system previously developed by the authors [30]. A BIM module is integrated to provide geometric and attribute data to both the human workers through the user interface and the Robot Operating System (ROS). After a robot generates work plans with information from the BIM, human co-workers supervise the robot's workflow (e.g., preview robot plans and monitor execution status) and make interventions (e.g., adjust installation target or request another trajectory plan) when necessary.

In addition, the as-built data collected by the robot during the construction process is sent to the BIM to reflect changes between the as-designed and the as-built workspace to be used for ensuing construction tasks, thereby closing the loop. The system supports seamless integration of various sources of information received from the BIM, human co-workers, and robot sensors. In parallel, the system processes the collected information into different forms that can be visualized and understood by humans, processed by robots for computing and control, and saved and presented in BIMs.

The BIM-driven HRCC processes involve five main steps, including (1) formulation of the BIM repository that supports robotic construction; (2) preparation of the construction site and the BIM for a specific construction activity; (3) automatic digital twin generation; (4) construction execution, including resolution of as-built / as-designed deviations; and (5) updating of the BIM repository with as-built construction data collected by the robot. The technical approach to enable these processes is discussed in Section 3.

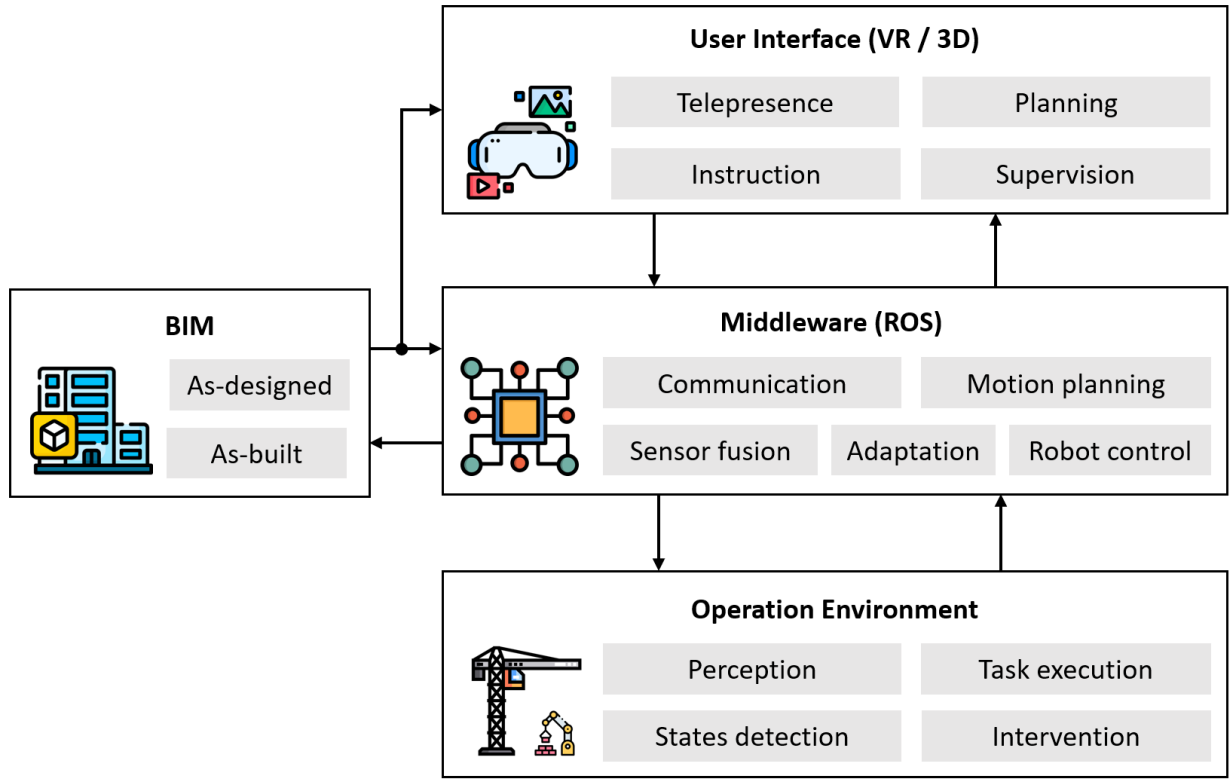

Figure 1: Closed-Loop Digital Twin Framework Overview

A drywall installation case study conducted on a large-scale physical Kuka industrial robotic arm is presented as a proof-of-concept implementation to explore the physical setup process and for system verification (Section 4). In addition, physical and simulation experiments involving block pick-and-place and drywall installation activities are carried out to evaluate system performance and validate the proposed deviation adaptation methods (Section 5). The proposed system not only extends the autonomy and accuracy of robotic construction but also offers the flexibility to overcome uncertainties in field construction work. The presented framework and workflow have the potential to be applied to a broad variety of construction tasks.

## 2   BACKGROUND

### 2.1   Human-Robot Collaboration

HRC aims to retain the robustness and adaptability of an automated system by combining robots' computational ability and durability with humans' creativity and flexibility [31,32]. HRC can be classified into two types, physical collaboration and contactless collaboration [33]. For physical HRC, the human co-worker intentionally makes physical contact with the robot or the object held by the robot to hand over or co-manipulate objects [34,35]. However, working alongside robots, especially construction robots that carry large and heavy components, poses high safety risks to humans. Therefore, contactless HRC has been used in many applications. It allows humans to guide robots with gestures [36], natural language [37,38], joysticks and haptic devices [39,40], Virtual Reality (VR) [13,41,42], and neural signals [43,44]. One main concern with contactless HRC, especially those from remote locations, is that humans' perception of the operational environment and the robot is limited. It is critical to provide humans with sufficient and accurate information in an effective way to enable efficient decision making. One of the popular approaches is to create a digital twin to provide real-time information about the robot and its operation environment to humans [45,46].

In prior work, the authors proposed an I2PL-DT system for human workers to remotely receive and visualize the updated state of a construction workspace [30]. The system consists of three modules, including an immersive Virtual Reality (VR) interface, middleware enabled by ROS, and Robot Operation Environment (ROE) (i.e., construction site, robot, and sensors). Human co-workers can visualize construction site and robot conditions in real-time and perform high-level task planning (e.g., indicating component installation sequence and positions). The high-level plan is then used to generate motion plans, which are processed into realistic animations in VR for the human co-worker to preview and evaluate. Upon approval, the robot executes the approved plan under human supervision. The work supports a collaboration paradigm where human workers perform high-level decision-making and supervision while robots undertake low-level motion planning and physical execution of the work.

Despite being a key component of the workflow proposed in this paper, the previously developed I2PL-DT system in and of itself has several limitations. First, the process of creating the digital twin for a construction task takes considerable effort. The immersive VR interface needs to be manually created by importing BIM data into VR, creating interactive game objects, and adding interactive functions (e.g., sending messages to ROS). Second, the human co-worker needs to specify the work plan for each component by indicating to the robot which component to pick up and the location to install it. Thus, substantial human effort is required for such step-by-step instructions. Lastly, since the human co-worker specifies the high-level task objectives by manipulating and placing virtual objects with controllers in VR, the accuracy is limited and the resulting work may not comply with typical construction work specifications and desired tolerances.

In order to overcome these critical and practical limitations, this paper integrates a BIM with the I2PL-DT system and proposes an automatic approach to creating digital twins. Figure 2 shows the elements of the updated and significantly expanded system presented in this paper and compares it to the previous I2PL-DT system. With the proposed approach, manual processes of creating the digital twin and placing each component step-by-step in VR can be avoided, which improves system autonomy and work accuracy as well as reduces human workers' workload. In the remainder of this section, the adoption of a BIM to support construction automation and robotics and the existing approaches to create digital twins for robotic applications are introduced and discussed.

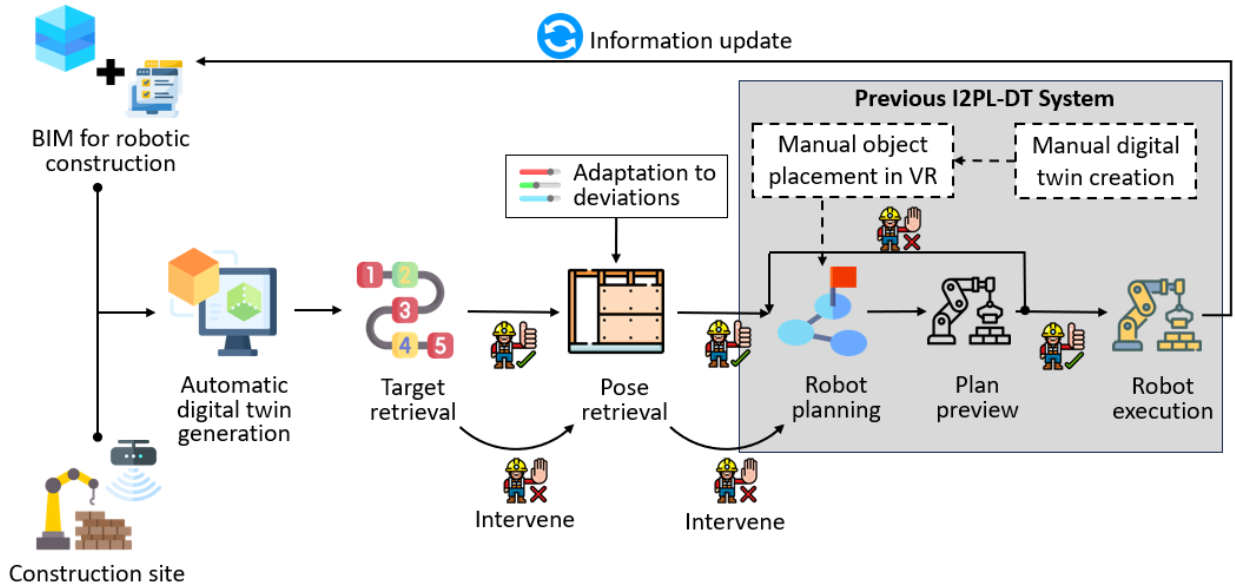

Figure 2: Elements of the Updated System

## 2.2 BIMs in Construction Automation and Robotics

BIMs have been widely adopted to promote automation throughout the life cycle of the Architecture, Engineering and Construction, and Facilities Management (AEC/FM) industry [47]. Recently, BIMs have been used to facilitate robotized construction in various ways. For example, BIMs can provide information to guide the off-site prefabrication process [48,49] and assist with object recognition for on-site assembly [50]. Layout information contained in the BIM is used to support robot indoor navigation tasks for building construction and maintenance [51–53]. Commercial mobile robots have been introduced to draw layouts on-site based on BIMs [54]. The geometric information contained in BIMs has also been used to facilitate 3D printing in construction [55,56].

     BIMs can also provide information to support robot motion planning. For example, in robotic brick assembly tasks, BIMs can provide position and orientation data for a robot to pick up and place materials as well as task sequencing data to control the robotic workflow [24,57]. IFC models have been converted to work with robotic simulation platforms for robot navigation [23] and to simulate the robotic wall frame assembly process [58]. In these existing studies, robots fully rely on the BIM for task planning. The construction site needs to conform to the exact BIM specifications for the robot to successfully perform construction work. However, considerable uncertainties exist on construction sites. Deviated components, moving workers and equipment, and stacked materials on-site may interrupt the BIM-generated robot motions, causing robots to stall while performing construction work on-site. [59] used a combination of the BIM and sensing information to generate adaptive robot motions. Despite such advances, the robot's adaptability is limited to a narrow set of situations and construction tasks [59].

     In summary, existing studies leveraging BIMs in robotics have three limitations. First, they lack generality to support various types of construction tasks. Second, they cannot handle uncertainties for field construction, which can go beyond the robot's adaptive capabilities and

interrupt robot activities generated by the BIM. Thus, human intervention is necessary in addition to the BIM for the success of robotic field construction. Third, as-built data collected by the robot during the construction process is useful for the ensuing construction, operation, and maintenance phases of the project; however, a closed loop for a BIM to both provide and collect construction information is missing in existing studies. Therefore, a general framework that supports different types of construction tasks and allows human intervention is necessary.

## 2.3 Digital Twins Creation in Robotic Applications

Digital twins can be used for visualizing and incorporating information from different resources, and they also support real-time communication and interaction [60,61]. Therefore, they are a promising candidate to integrate BIMs with HRCC. Based on the demand for different application purposes, digital twins can be created with various approaches. One of the most popular methods in the AEC/FM industry is to use 3D point clouds. The environment is captured with laser scanners or depth cameras as 3D point clouds [62–64]. Such systems can comprehensively capture the environment in real-time, theoretically allowing for continuous updates of the digital twin to reflect the evolving environment at any point in time. However, the transmission of large-size point cloud data is computationally expensive and is subject to delays. Furthermore, subsequent steps of processing and registering the point clouds, object detection, or 3D reconstruction are usually needed to achieve the required functions [65,66].

[67] developed the VITASCOPE language to integrate simulation models, hardware control, and real-time construction data for dynamic construction scenario visualization at the operation level. [68] and [69] created digital twins of a robot to program, control, and visualize robot motions. Joint state data are exchanged between the physical and virtual robots in real-time. In order to reduce the computation load while allowing real-time visualization of the construction environment, [30] created a digital twin using a combination of a BIM, 3D meshes of as-built structures, and point clouds. However, these approaches require manually importing models or enabling the functionality of the digital twin. As a result, it interrupts the automated workflow of HRCC initiation, which can potentially hinder the widespread application of digital twins in HRCC. Therefore, an approach to automatically generate digital twins that are equipped with pertinent functions (e.g., target object selection, communication between modules) is necessary to promote and extend the autonomy of the HRCC workflow.

## 2.4 Research Objectives

To address the abovementioned research gaps, a workflow with the following characteristics is needed to effectively support HRCC. First, instead of parsing specific information from BIM repositories to automate one type of construction work, a general framework that supports different types of tasks and components is needed. Second, the workflow should automatically interface with different construction tasks without additional programming or development effort that cannot be performed by construction workers without related expertise. Third, robots should have the ability to handle significant uncertainties on construction sites while allowing human intervention to resolve cases that extend beyond the robots' capabilities. Lastly, the workflow should record the construction-related data collected by the robot sensors during the HRCC process for future reference to enable loop closure. With these objectives, the following section presents a closed-loop generalizable framework integrating a BIM and HRCC, which supports automated collaborative workflows and can overcome uncertainties in field construction work.

## 3 TECHNICAL APPROACH

A BIM-driven HRCC workflow enabled by closed-loop digital twins is proposed in this study. The BIM provides data (e.g., component geometry, position, type) to both the robot and the human co-workers by compiling and sending messages to ROS and Unity (where the interface is developed). The robot then generates the work plan based on the BIM and adapts the plan according to the as-built circumstances detected by its sensors. For example, during drywall installation, the robot retrieves the name and target position of the next panel to install from the BIM; it then adjusts the installation position of the panel based on how the wall frame has actually been built as detected by its camera. Next, human co-workers supervise the robotic construction process (e.g., by evaluating and approving the robot plan) and intervene to adjust the high-level task plan (e.g., by adjusting the target installation sequence and pose proposed by the robot) when necessary.

An overview of the BIM-driven HRCC process is shown in Figure 3. First, the BIM data accessible by both human co-workers and robots needs to be created. Next, before construction starts, human workers need to prepare physically on the construction site (e.g., materials staging) and digitally in the BIM (e.g., task designation). Then, the interactive digital twin is generated, followed by the construction execution. Lastly, after certain construction tasks are finished, the BIM repository is updated with the latest as-built data. The remaining part of this section introduces the system design and the technical approach to enable this process.

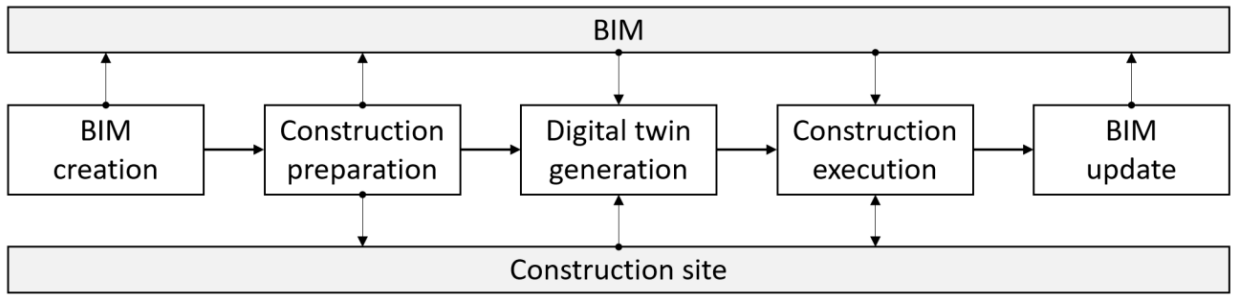

Figure 3: BIM-driven HRCC Process

### 3.1 System Design

#### 3.1.1 BIMs for Robotic Construction

BIMs created for traditional construction methods are typically not compatible with robotic construction [70]. Some additional elements are necessary for a BIM to better support human-robot collaborative task planning:

- ***Shop drawing-level geometry.*** Shop drawings precede detailed work plans and contain information needed for fabrication, assembly, and erection [71]. For example, in a robotic drywall installation task, shop drawings for drywall panels are needed so the robot knows where each panel should be located. However, shop drawing-level details are not needed for every component in BIM and are dependent on the construction plan. For example, an object that is prefabricated off-site or erected on-site as a whole can be represented in the BIM as one single component.

- ***Construction sequence.*** Detailed construction sequence information contained in the BIM is crucial to automate the construction process. The construction sequence information should be provided in correspondence with the shop drawing-level geometry, specifying the order of installation of components.

- *Component relationships.* Relationships between components can affect robot planning, especially when discrepancies exist and the robot needs to adapt its plan. Even though some component arrangements visually look the same, they might indicate different relationships (i.e., topologies) which require different construction plans. As a result, clearly defined component relationships in the BIM are necessary for the adaptive planning of the human-robot work team.

- *Component layers.* A unified predefined layer structure in the BIM not only helps organize components into different groups for user comprehension but also supports the development of interfaces that can quickly connect to different BIM projects and automate related processes in the project life cycle, such as construction. Therefore, a BIM layer structure is proposed to facilitate the automatic generation of interactive digital twins and the HRCC process, as shown in Table 1.

Table 1 BIM Layer Structure for Robotic Construction

| Layer Name | Description |
| --- | --- |
| Target | Physical objects (e.g., timber) or virtual indicators (e.g., fastener locations) the robot needs to install or operate for the current construction task. |
| As-Built | Components that have already been built. If they are inside the workspace, they need to be considered for collision avoidance. They update with the construction progress or as the robot senses the environment. |
| Materials | Construction materials staged on-site. If they are inside the workspace, they need to be considered for collision avoidance and may be updated during the construction process. |
| As-Designed | The original design of the components. The information is used by human co-workers and robots to understand as-designed versus as-built deviations and develop plans accordingly. |
| Virtual Collision | The space that is not physically occupied but the robot needs to avoid during the movement (e.g., a safety laser curtain that marks the edge of the work zone). The robot considers it for collision detection during motion planning. |

- *Robot operation support.* Adding information to support robot operation in the BIM can facilitate HRCC work. The information added depends on the task type and robot intelligence level. For example, robots can determine how to grip components by visually detecting component geometry in some cases, but for some components with irregular shapes, external guidance can increase the success rate of component gripping.

The creation and storage of these elements depend on the BIM platform selected. In this study, Rhino is used as the BIM platform [72]. The shop-drawing geometry, component name, and layer information are stored with the software's default field. Additional attribute data are stored as "Attribute User Text" in Rhino. For example, the component relationships are denoted by the "Parent" attribute of the components, while the sequence data are held in the "Sequence" attribute. To facilitate robot gripping, a Python script automatically calculates the gripping reference points from the component's vertices and writes the data to the corresponding component attribute. This geometry, layer, and attribute data are extractable through scripts and can be communicated to ROS and Unity.

It should be noted that, although being manually created in this study, some of the abovementioned information has a high potential to be automatically generated by the computer or the robot. For example, [73] proposed an approach for automatically generating steel erection sequences and [74] used convolutional neural network and large-scale robot grasping experiments to generate robot grasping plans. [75] used computational design to generate the cutting planes, gripping planes, and connections for off-site frame prefabrication. This research focuses on how to leverage such information for HRCC; manual addition is thus sufficient for this study.

### 3.1.2 BIM-Driven HRCC Framework Design

Five elements are included in the proposed framework for BIM-driven HRCC: 1) the BIM that provides and saves data about the construction project; 2) the Graphical User Interface (GUI) that supports both immersive VR and 3D options for human workers to interact with the robot; 3) ROS as the middleware for communication and the central unit for data processing, computing, and construction work process and physical robot control; 4) ROE, which is the construction site that includes robots, sensors, and materials; and 5) human workers who supervise the construction process and intervene when necessary. In this study, Rhino 7 is used as the BIM platform, and the GUI is developed in Unity with Oculus Rift S as the headset for the immersive VR option.

The information flow among system elements at different stages is shown in Figure 4. To prepare for construction, setup activities are needed both physically on the construction site and digitally in the BIM. In general, workers need to start the sensors that are in use (e.g., cameras, LiDARs) on the robot or construction site for the robot to perceive the environment. They also need to place the required construction materials in the robot workspace for the robot to reach and manipulate. Lastly, workers need to assign tasks to the robot in the BIM (e.g., by selecting some drywall panels on the frame and placing them onto the "Target" layer). They should also adjust components' status in the BIM according to the task scope (e.g., marking components that are outside the robot workspace as "unrelated" by changing the user attribute data). Considering that a BIM interface is intuitive and generally familiar to construction personnel, it is typically easier for workers to indicate task scope in a BIM rather than directly indicate this to the robot in ROS.

When construction starts, information is taken from both the BIM and the construction site to generate the I2PL-DT. An automatic digital twin generation approach is proposed, which is introduced in Section 3.2.1. During the construction process, human co-workers collaborate with the robot through the digital twin system, which integrates information from the BIM and the construction site. The detailed approach to updating the digital twin and enabling HRC through bi-directional communication is discussed in Section 3.2.2. Section 3.2.3 introduces how the system addresses the challenges caused by component deviations on-site. The robot can suggest solutions to adapt to deviations to human co-workers for approval. Occasionally, human interventions (e.g., replacing a component with one of a different shape) are required to resolve the deviations.

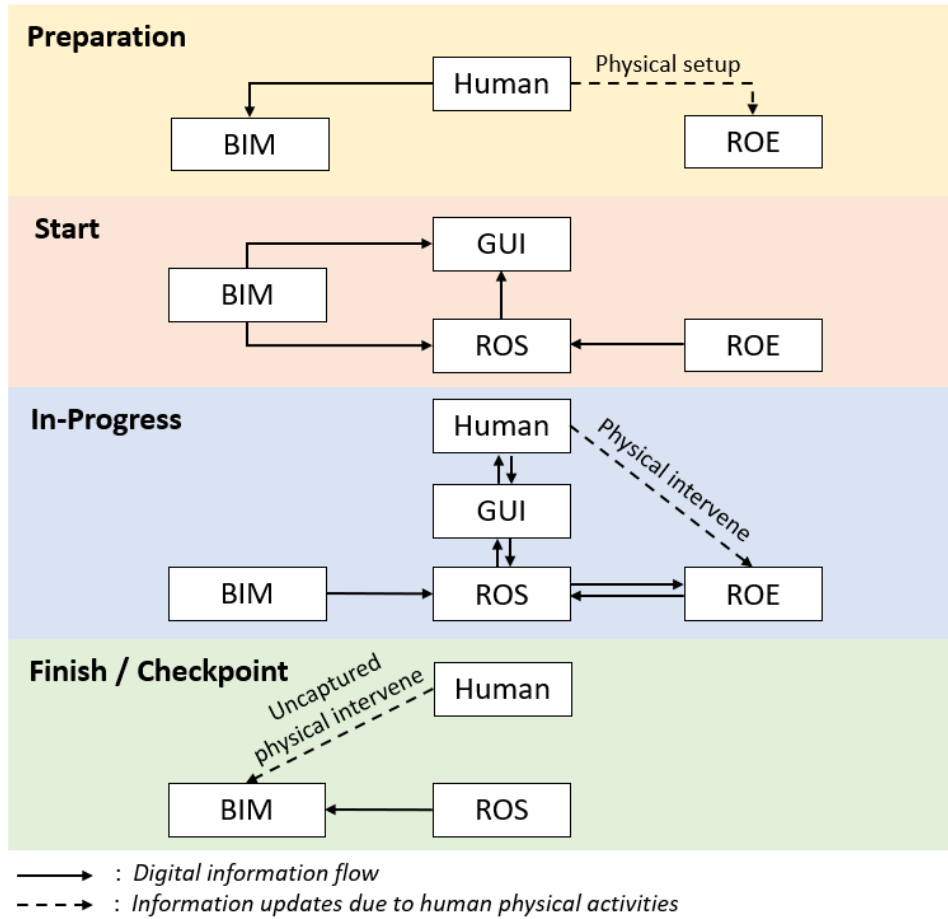

Figure 4: System Information Flow at Different Stages

When construction completes or reaches a certain checkpoint, the ROS environment that contains the related construction site information at the time sends the information to update the BIM repository and records data in the BIM for future reference. The technical approach to updating the BIM with construction data is introduced in Section 3.2.4. It must be noted that even though this study chooses to update the BIM at certain time points, the framework also allows the BIM repository to be continuously updated as construction progresses. Lastly, Section 3.2.5 explores the deployment of the physical system to collect and integrate data from the physical world and to control the physical industrial robot.

## 3.2  Technical Approach for System Implementation

### 3.2.1  *Automatic Digital Twin Generation*

To initiate construction, it is essential to generate the digital twins that integrate visualization, interaction, and computation functionalities. Our system incorporates two digital twins. One is the interaction module with the GUI developed in Unity, which facilitates visualization, supervision, and intervention. The other is the computation module within ROS that manages the construction workflow (e.g., installation sequence), adapts to as-built conditions, and generates task plans (e.g., collision-free motion plans). The generation of these digital twins takes advantage of the geometry and attribute data of components from the BIM. The coordinates of the digital twin world and the BIM are aligned using the robot's base frame as the origin point. Sensor data are also transformed into the robot base frame for integration into the digital twins. For example, for the camera sensor mounted on the robot in the experiments of this study, its pose relative to the robot link it is attached to is calculated through hand-eye calibration [76]. The relative pose is further transformed to the robot base frame using forward kinematics according to the current robot joint states. The automated generation process of the digital twin system is shown in Figure 5.

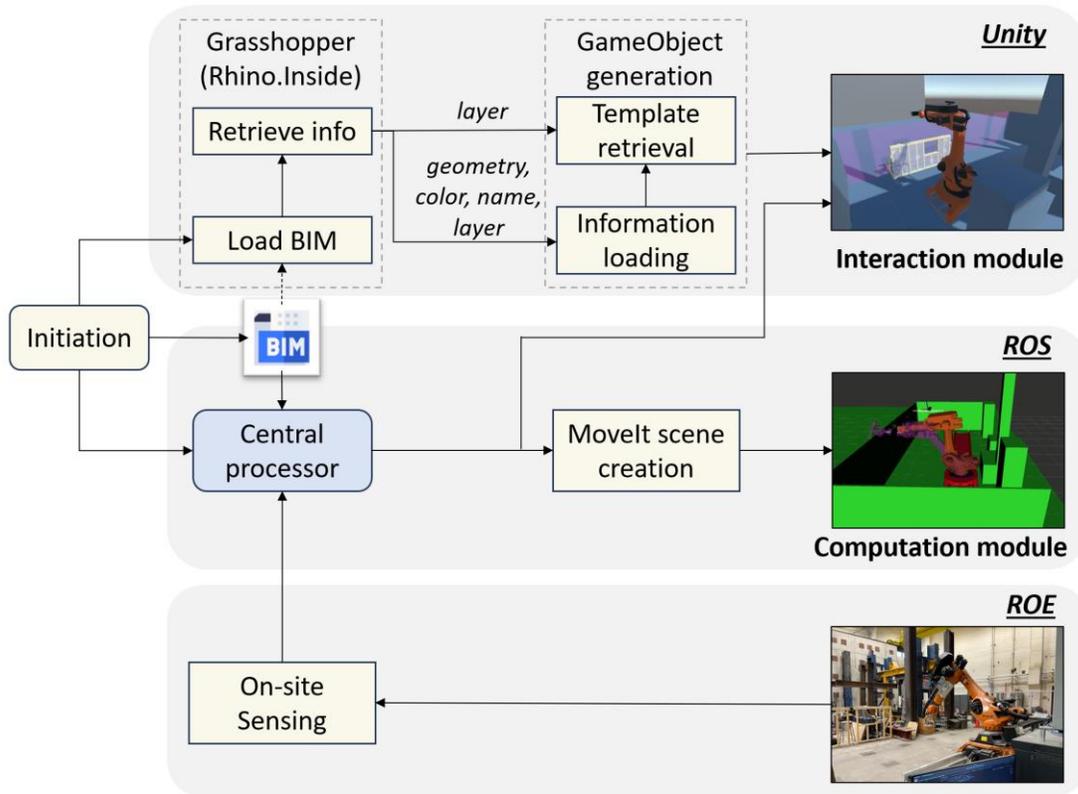

Figure 5: Automatic Digital Twins Generation Process

The computation module is developed in ROS. Geometries of task-related components in the BIM are extracted as meshes, converted into ROS messages by a script in Rhino, and subsequently sent to ROS. Attribute information, including names, layers, and customized properties designated as Attribute User Text (discussed in Section 3.1.1), are also sent to the computation module in the form of ROS messages. As-built data, including the locations of as-built components and materials, are loaded into the module when they are detected at the construction site. During the construction process, the system retrieves these data for task and motion planning. For example, the system queries the parent of the target and whether it is deviated to determine if adjustments are needed to install the target object.

Motion planning in the computation module is facilitated by the MoveIt planning framework [77]. The planning scene of the MoveIt framework contains a virtual robot model that replicates the actual robot using the URDF descriptions and synchronizes with the actual robot's states by subscribing to the joint_states topic. The planning scene also incorporates environmental representations, where meshes of task-related components on the "As-Built", "Materials", and "Virtual Collision" layers in the BIM are spawned at their corresponding locations as collision objects. The robot will use its camera to detect as-built components and construction materials in the workspace and subsequently update the collision information in the planning scene. If a component has been constructed but its as-built model is not yet present in the BIM (not detected by the robot), its corresponding as-designed model will first be generated and later replaced by the as-built model once detected (e.g., through fiducial markers). Detected construction materials are similarly updated in the digital twin. These collision objects are used by the robot to plan collision-free paths during motion planning. When holding a material, the corresponding collision object is attached to the robot arm to ensure it is considered during motion planning.

The interactive digital twin module in Unity is generated by directly connecting Rhino and Unity through Rhino.Inside [78], which is an open-source add-in that enables other applications (e.g., Unity) in the Windows Operating System (OS) to run Rhino and Grasshopper (a visual programming tool integrated with Rhino [79]) projects. The highlight of this process is that instead of manually importing a BIM and creating functions in Unity for different construction projects, a Unity program template is developed. It contains template models with functions. When it receives information from other modules (e.g., a BIM), Unity can quickly generate interactive game objects using these template models according to the object layer. For example, when it receives an object assigned to the "Target" layer from a BIM, the program instantiates a game object with the target-type template. This game object is equipped with functionalities for user selection and pose manipulation. Meanwhile, the mesh geometry received from the BIM is loaded onto the game object for visualization, and the game object attributes such as name, layer, and material are set accordingly. As a result, the template Unity program can automatically generate interactive digital twin interfaces for different BIM projects.

At the initial state, Unity contains 1) light and camera systems for human partners to visualize objects in the game interface; 2) event systems to capture user input and enable interaction (e.g., selection, movement); 3) virtual robot models generated from the Unified Robot Description Format (URDF) files and meshes from ROS. One virtual robot is synchronized with the actual robot for supervision. The other one is used to evaluate robot motion plans and only appears when the human workers preview plans; 4) ROS connectors to exchange data with ROS through Rosbridge using the ROS# library [80]; 5) a game object to run Rhino and Grasshopper projects to retrieve geometry and attribute (e.g., layer, name, type) data from the BIM; 6) model templates that can automatically generate interactive game objects with the information received; 7) interface templates containing text instructions and buttons that appear at specific times to prompt and receive user input; and 8) a virtual billboard to show messages.

As the system is initiated, Rhino and Grasshopper applications start with the Unity program through Rhino.Inside [78]. A script developed in Grasshopper then loads the BIM into Rhino, retrieves information (i.e., geometry, color, name, layer) of objects from the BIM, and sends it to Unity. As Unity receives the information, model templates can create game objects with the same names, colors, and geometry as what they received from the BIM. The generated game objects are automatically placed onto the corresponding layers in the GUI. Based on the object layer, scripts that contain different functions are embedded in the model templates and attached to the object, thereby creating different interaction patterns. At the same time, the communication between Unity and ROS is established through the Rosbridge WebSocket server [81], which enables the seamless interaction between the human co-worker and the robot during the construction process (Section 3.2.2). Lastly, the robot moves to scan materials and as-built components in the workspace. As-built components and materials detected are instantiated at their corresponding locations in Unity to reflect the most up-to-date construction environment to the human co-worker. With these steps accomplished, the digital twin systems are completely generated and construction can start.

### 3.2.2 Digital Twin Update for Collaborative Construction Implementation

During the construction process, the seamless integration of the human co-worker and the robotic system is achieved with communications through Rosbridge [81], with ROS serving as the middleware. Data exchange occurs at various frequencies throughout the construction process. For example, joint states are written onto the physical robot at a fixed frequency of 250 Hz. In contrast, certain data are only published when specific events are triggered (e.g., the human co-worker makes a selection from the GUI). The HRCC process is shown in Figure 6. Human operations and decisions are shown in orange, and blue elements show processes performed by the robot. The robot and its human co-workers interact through the GUI. The overall workflow is that the robot shows information and decisions intuitively in the GUI for its human co-workers to visualize. Then, the GUI detects the human co-workers' decisions and operations through their input and sends the information to the robot. During the interaction process, corresponding interfaces are generated from the interface templates to prompt the human co-workers on the current process and get inputs. The templates contain corresponding functions (e.g., sending messages to the robot on certain ROS topic channels when the user clicks certain buttons) and are automatically connected to corresponding scene objects when being generated.

After generating the digital twins, the robot retrieves the next target in the construction sequence from the BIM and highlights the target object in the GUI in semi-transparent green. Meanwhile, it prompts the human co-worker to confirm the target by instantiating a UI in Unity with the natural language text "*Want to install this?*", where the co-worker can respond by choosing either the "Yes" or "No" button. If "No" is selected, the human co-worker can select another target through mouse clicks on a computer or ray casting in VR. After a target is confirmed, the robot checks whether there are deviations in the workspace that affect the operation to achieve the target as originally designed (e.g., a deviated component occupying the target space will collide with the target if following the original design). If there is a deviation, the robot will propose suggestions to adapt (see Section 3.2.3 for the adaptation approach). Note that the adaptation decision is sometimes based on the robot's suggested solutions and sometimes based on human improvisations, thus the "adaptation" block appears in two colors. If no deviation exists, it will go ahead with the original installation pose from the BIM. The human co-worker can choose to directly accept the robot's suggestion (e.g., an adjusted pose to install the target). There might be situations where the robot cannot properly adapt, so the system also allows the human co-worker to improvise an adaptive plan (e.g., move the target to indicate a desired installation pose in GUI or directly install the component by themselves).

Next, based on the confirmed plan, the robot generates a collision-free motion plan to achieve the target. When the human co-worker requests to preview the plan, a virtual robot will appear in the GUI and demonstrate the motion plan as an animation. The animation includes the robot movement during the whole manipulation process and the material movement if it is being held by and moving with the robot. This is enabled through a virtual joint state publisher in ROS that extracts and publishes the joint states in the generated motion plan at a given frequency. A virtual robot emulator in Unity subscribes to the topic that publishes the virtual joint states and moves with the states received, thereby allowing previewing of the motion plan as an animation. Such previewing processes enable the human co-worker to gain insights into the consequences of the robot operation and make better decisions [82]. If the human co-worker is not satisfied with the motion plan, the robot will generate another plan with MoveIt and demonstrate it in the GUI until the human co-worker finds an acceptable plan. After a motion plan is accepted, the actual physical robot executes the approved plan. A detailed introduction of the processes to preview, execute, and supervise the robot motions can be found in the authors' previous work [30].

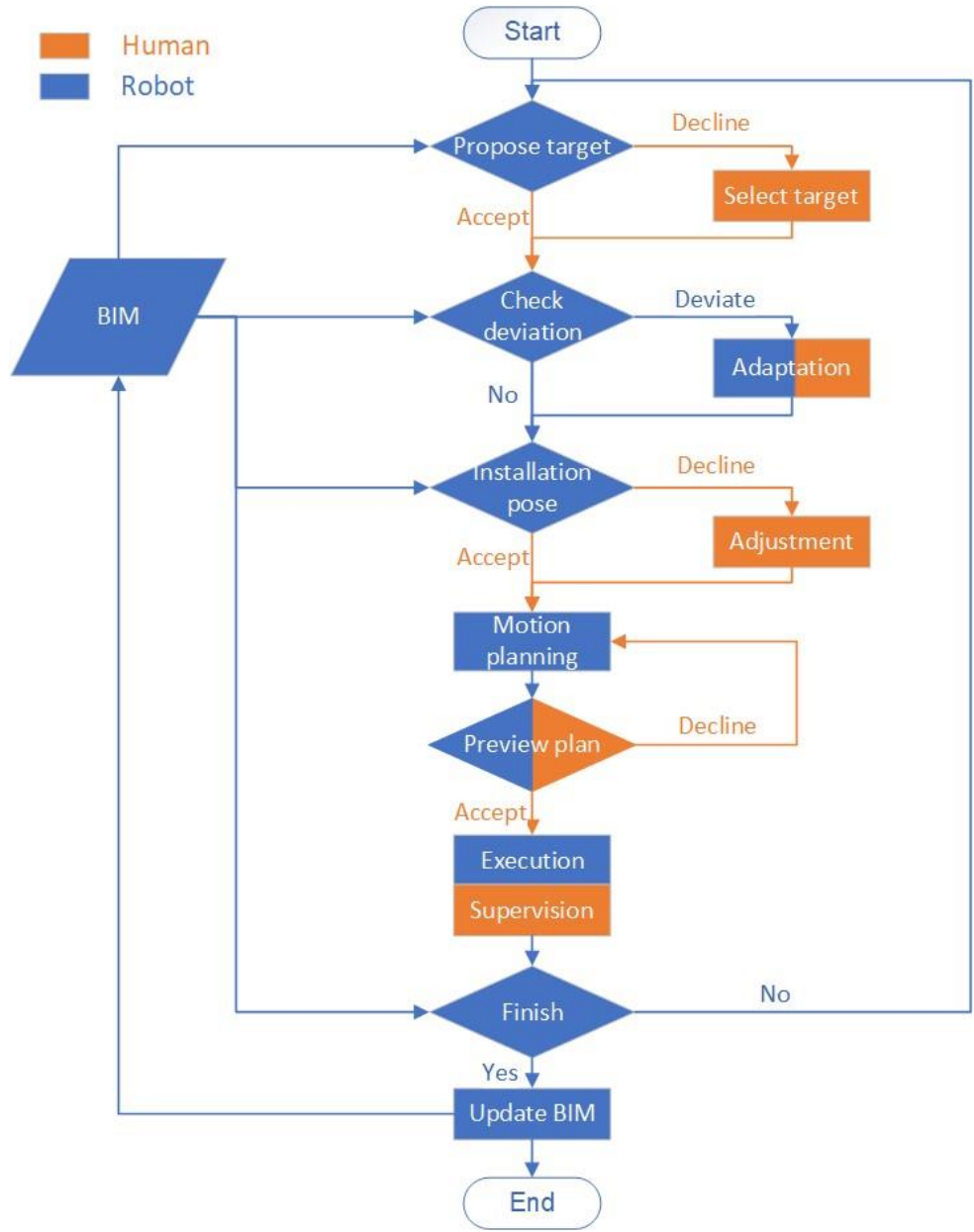

Figure 6: Collaborative Construction Process Flow Chart

    Since a robot model in GUI is synchronized with the actual robot by subscribing to its joint states, the human co-worker can supervise the robot execution states through this synchronized robot. They can also understand the robot's cognitive status (e.g., calculating the motion plan) through the virtual billboard in the GUI. It should be noted that certain steps of processing and robot operation can be skipped, depending on the type of human intervention. The technical approaches for establishing connections within the digital twin system and achieving motion

planning, plan preview, and execution and supervision functions are discussed in [30]. After a target is achieved, the system will check the next target in the sequence from the BIM and go through the process again. If no target is left in the queue, the assigned task is considered to be finished.

### 3.2.3 Deviation Adaptation

This study develops methods to enable robots to provide suggestions for adapting to two types of deviations commonly found in construction assembly tasks. The process is supervised by the robot's human co-worker, and if needed, the human can intervene to adjust the robot's suggested solution or improvise a different adaptation strategy.

*__Parent deviation:__* The parent of the target to be installed is built with deviations from its design. Since the target needs to be connected to its parent, the installation pose of the target should be adjusted accordingly. Two sources of transformation information are used to address the deviation. The robot receives the as-designed position and orientation of the parent object from the BIM through Rosbridge at the beginning of the construction, which is then converted into a transformation matrix $T_D$. The as-built transformation of the parent object $T_B$ is detected by the robot with the camera. After both transformations are received, the robot first calculates the design-built deviation $T_D^B$ of the parent using $T_D^B = T_B T_D^{-1}$. Next, it calculates the suggested transformation matrix $T_t$ to install a target $t$ using $T_t = T_D^B D_t$, where $D_t$ is the as-designed transformation of the target coming from BIM. Lastly, the installation transformation matrix $T_t$ is converted into an installation pose $P_t$ with position and orientation information for robot planning and operation.

*__Nearby object deviation:__* When certain objects near the target are built with deviation, they might occupy the originally planned space of the target. Before installation, the robot checks whether the planned installation place collides with any as-built objects. This process is enabled by the collider functions in Unity. If no collision is detected, the robot will plan to go ahead with the original plan by default. Otherwise, the robot will provide suggestions to offset the target installation pose to avoid collision based on the deviation of the object that conflicts with the target object.

*__Human supervision and intervention:__* The robot's suggested solution always requires human approval before execution to prioritize human preferences. The suggested installation pose is communicated to human co-workers by instantiating a new object with the same geometry as the target object highlighted in semi-transparent red material in the digital twin interface. Meanwhile, an interface template is instantiated asking "*Robot suggest install it here. Do you accept?*". The human can choose to take the robot's suggested solution by clicking on the "Yes" button. However, due to the possible uncertainties and complexities of construction work, the deviation type may not belong to the two situations considered above or the solution suggested by the robot may not be optimal, especially for the nearby object deviation case where situations vary. For example, offsetting one component may affect the installation of subsequent targets, causing construction plan changes for several targets. In this case, the human co-worker may prefer to manually replace the target with one in a different shape (e.g., a smaller piece). Even if the deviation does not introduce collisions (object deviates towards other sides), the human co-worker may still want to adjust the target workpiece to make it stay together with the objects nearby. In these cases, the human co-worker will reject the robot's suggested solution and manually intervene.

The system affords two operations to facilitate human intervention. First, they can manually adjust the position and orientation of the target installation pose through sliders in the interface. Otherwise, they can skip the robotic installation in the digital twin and manually adjust and install the target. In this case, they need to manually record the information in the BIM to make it available for future use.

### 3.2.4 BIM Information Update

When an assigned construction task finishes or reaches a certain checkpoint, three sources of data tracked by ROS can be sent to the BIM via Rosbridge using the COMPAS library [83]. Three sources of data are saved in the BIM repository for future reference. The workspace sensing data as the robot scans the environment and the robotic construction data reflected by robot states are saved onto the "As-Built" layer. The temporary material data (i.e., on-site construction material type, number, and locations) are saved onto the "Materials" layer. It is inferred from the start state (e.g., materials originally prepared) and the construction process (e.g., how many materials are used).

To update the BIM, the pose and type (applies to Materials only) data of these components in the ROS computation module are sent to the BIM. Then, new components are instantiated in the BIM at the given poses, with their geometries retrieved from corresponding components previously existing on the other layers in the BIM. The Python script in Rhino assigns names and layers to the instantiated components and generates customized attribute data (e.g., component poses) as Attribute User Text. Meanwhile, the components on the "Target" layer that are installed by the robot are switched to the "As-Designed" layer. In the ensuing steps, the repository will continue updating itself with incoming sensing data and completed construction work.

### 3.2.5 Physical System Deployment

The physical portion of the system is deployed on a large-scale Kuka industrial robotic arm in a research laboratory setting designed to mimic a construction site. The industrial robotic arm is selected because it has a relatively large payload to manipulate heavy construction components and higher flexibility to perform various construction tasks with different tools. The assumption is that the robot has a relatively static workspace (e.g., an area in a room) for each construction task. Other workers and equipment conduct construction activities and move outside the robot workspace. For safety reasons, if they get into the robot workspace during the robotic construction process, the robot will stop moving until safety is confirmed by its human co-worker.

The system involves several devices connected to a Local Area Network (LAN). Devices can communicate with each other through wired connections or wirelessly. The device communication and robot control processes are shown in Figure 7. The interfaces that the human co-worker directly uses, including the GUI in Unity and the BIM in Rhino, need to run on a computer with Windows OS. The sensors are connected to portable microcontrollers (e.g., Raspberry Pi). Both the Windows computer and microcontrollers can communicate with ROS wirelessly through a router on the LAN via ROS messages. ROS runs on a computer with Ubuntu OS that is connected to the robot embedded PC through an Ethernet cable. The computing core in ROS sends the joint states to the Automation Device Specification (ADS) interface of the programmable logic controller (PLC) [68]. ADS is an interface layer of Twincat PLC that allows commands and data exchange between different software modules [84]. The joint states data is received by the PLC and is then sent to the Kuka Robot Sensor Interface (RSI) to control the robot [68].

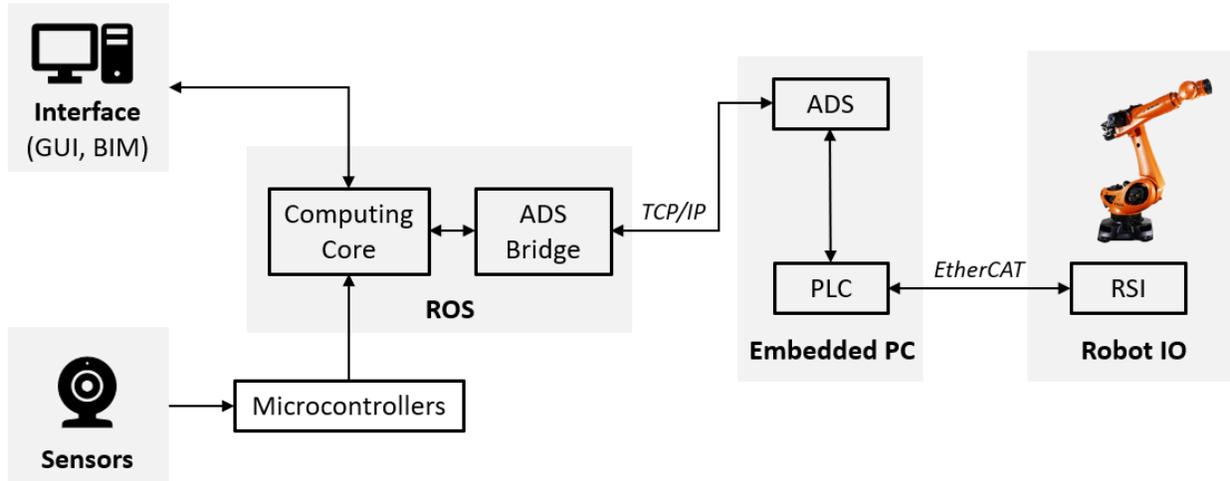

Figure 7: Physical Robot System Framework

## 4 SYSTEM VERIFICATION AND CASE STUDY

In order to verify the proposed system and explore the setup necessary for its physical deployment, a drywall installation case study is conducted in a research laboratory, which is set up to mimic a construction site, as a proof-of-concept implementation. Drywall installation is one of the most prevalent construction activities. It is also a representative example of large-scale object manipulation and pick-and-place operation, which is a common scenario in construction and comprises the elemental motions for a variety of construction tasks on structures (e.g., framing), surface (e.g., ceiling tile installation), and systems (e.g., ductwork installation) [37,85]. In this context, the wall frame itself is prefabricated without deviation but is purposefully installed at a deviated pose on-site to simulate the case of parent deviation. Four drywall panels in two different shapes need to be installed onto the wall frame, simulating scenarios such as windows, doors, or room corners that require varied panel sizes beyond the standard dimensions. The remaining part of this section introduces the physical and software setup and describes the HRCC process to perform the drywall installation task in detail.

### 4.1 Physical Setup

The robot used for the case study is a 6 DOF Kuka KR120 industrial robotic arm that has a 120kg payload and a 2.7m reaching range. By mounting it onto Kuka KL4000 Linear Unit, its base can move 4.5m linearly, which adds one DOF to the robot and significantly increases the robot's physical reach. Therefore, the robot has the capability to manipulate a regular-sized drywall panel. The robot workspace is shown in Figure 8. A safety gate is used to mark the robot workspace and a laser curtain is installed on the safety gate to prevent other workers and equipment from entering the robot workspace.

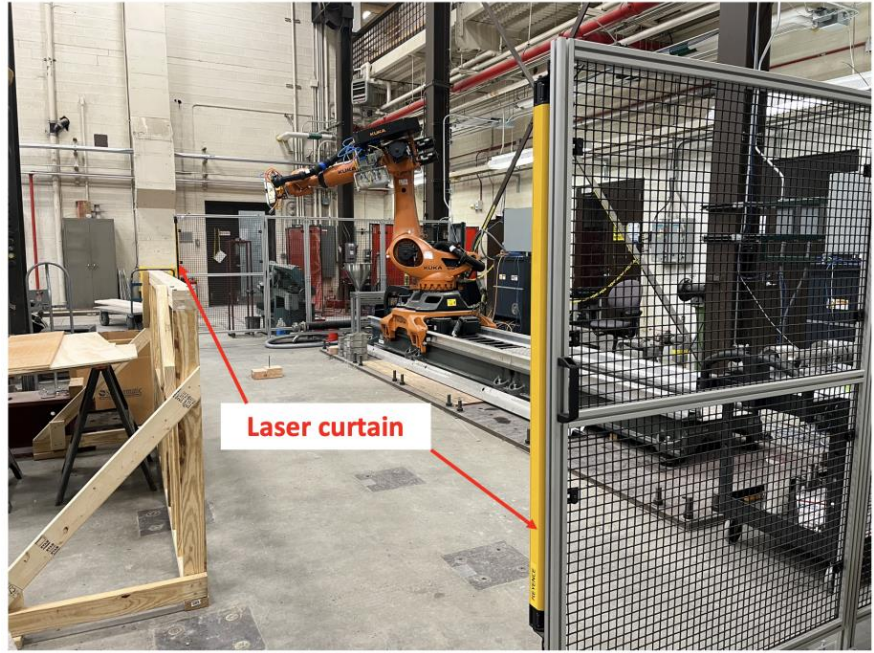
Figure 8 Robot Workspace

Since the goal of the case study is to verify the capability of the system framework and functions, the experiment is conducted on a 1:2 scale. A wall frame 4 feet (1.2192 m) tall by 8 feet (2.2384 m) long with a window area is built with studs on the back side to ensure stability (Figure 9). To expand the workspace, part of the frame support is placed outside the gate, but it does not break the operation because it is lower than the bottom of the curtain. The robot needs to install three larger drywall panels of 2 feet (0.6096m) by 4 feet (1.2192m) and one smaller panel of 2 feet (0.6096m) by 2 feet (0.6096m) onto the frame. A cubic handle is attached to each panel for the robot to grip. A pneumatic gripper is designed and connected to the robot with a tool changer and a connection plate (Figure 10). The jaws of the gripper are made with slopes to clutch the cubic handle on the drywall panel. Stabilizers are installed to ensure that the drywall panel can fully contact the gripper to avoid torque and shaking during manipulation. Rubber pads are used to add friction between the gripper and the cubic handle to prevent slippage.

Given that only one robotic arm is deployed, for demonstration and experiment repeatability, Medium-Density Fiber (MDF) boards with higher durability are used as drywalls. Magnets are used to attach the panels to the wall frame after the robot releases the panels. For actual construction work, the panels could be fixed (typically with screws) onto the wall frame, potentially by a human worker or another robot. An RGB camera is fixed onto the gripper. It is connected to ROS running on a Raspberry Pi microcontroller powered by a portable battery. Raspberry Pi can send the camera sensing data as ROS messages wirelessly to the ROS master running on the Linux machine. AprilTag fiducial markers are positioned near the wall frame and on the panels, and can be localized with the RGB camera [86]. The markers store information about the component type and the offset from the marker to the component's origin to help identify the components and their 6DOF pose. The type and offset information can be easily modified by updating the configuration file to accommodate multiple tasks.

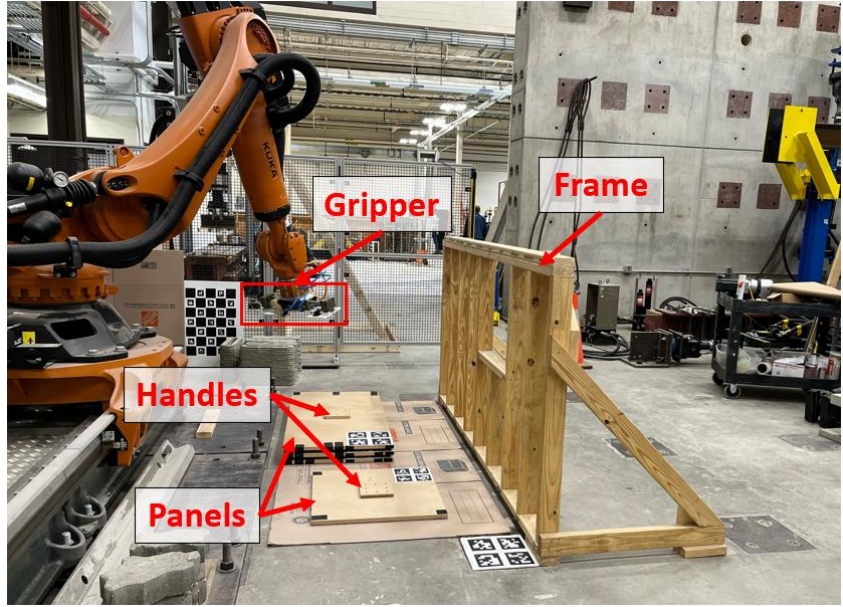

Figure 9: Wall Frame and Drywall Panels Used in Case Study

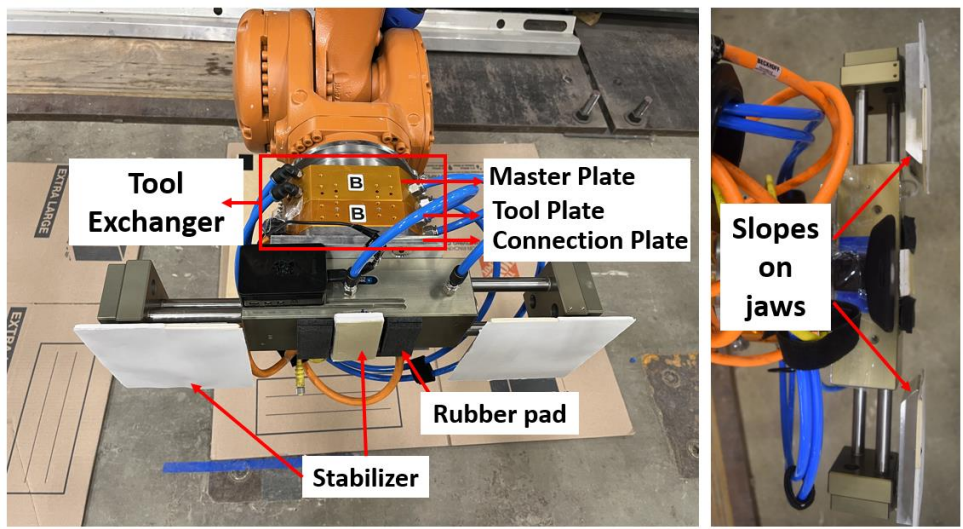

Figure 10: Gripper Design

## 4.2 Preparation of the BIM

Figure 11 shows a screenshot of the BIM used for the drywall installation task. The surrounding wall of the laboratory is set to be transparent grey to make it easier to visualize the robot workspace. Shop drawings for drywall installation are shown at the right bottom of the figure. The BIM indicates how the panels are designed to be installed. In this drywall installation task, these panels are specified as targets. The laser curtain is the plane that lasers come through which does not physically exist. However, interruption of the laser will cause the robot to stop for safety reasons, so the robot should not get into the curtain during operation. Thus, the laser curtain is considered a collision object during robot motion planning. The BIM components and their corresponding layers are shown in Table 2. Before construction, the frame has already been installed and materials are prepared but their poses are unknown and need to be detected by the robot.

Table 2 The BIM Components List

| **Components** | **Layers** |
|---|---|
| Workspace surroundings (Blue) | As-Built |
| Frame (Design) (Yellow) | As-Designed |
| Target panels on frame (Design) | Target |
| Laboratory walls, breams, and columns | As-Built |
| Laser curtain (Virtual) | Virtual Collision |

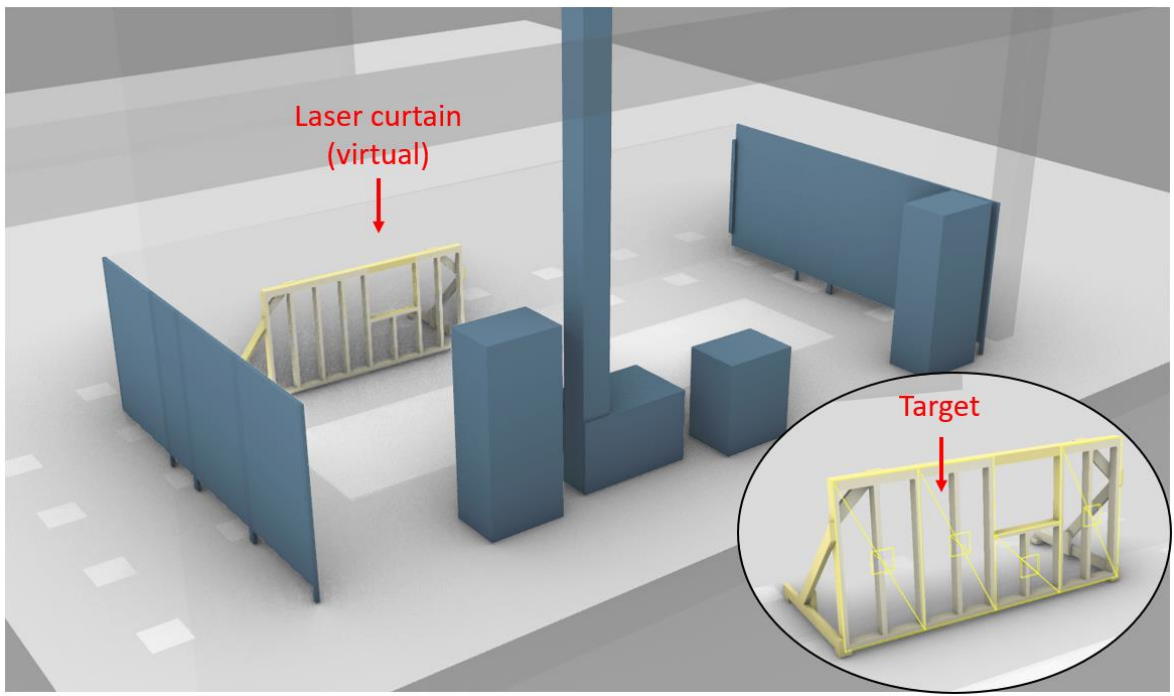

Figure 11: The BIM for Drywall Installation

The BIM contains the component attribute needed for robot processing and construction, such as its identifier, layer, its relationship with other components, whether it is related to the current task, construction sequence, and type (e.g., large or small) (Figure 12). For manipulable components, how the robot should grip the component is also indicated (e.g., at its center with orientation perpendicular to its largest surface), which is also used as the indicator of robot picking and installation pose. In this study, the program automatically calculates pose indicators using the centroid of the Rhino object. Otherwise, pose indicators can be automatically generated with computational design or human workers can indicate them by selecting points and directions in Rhino.

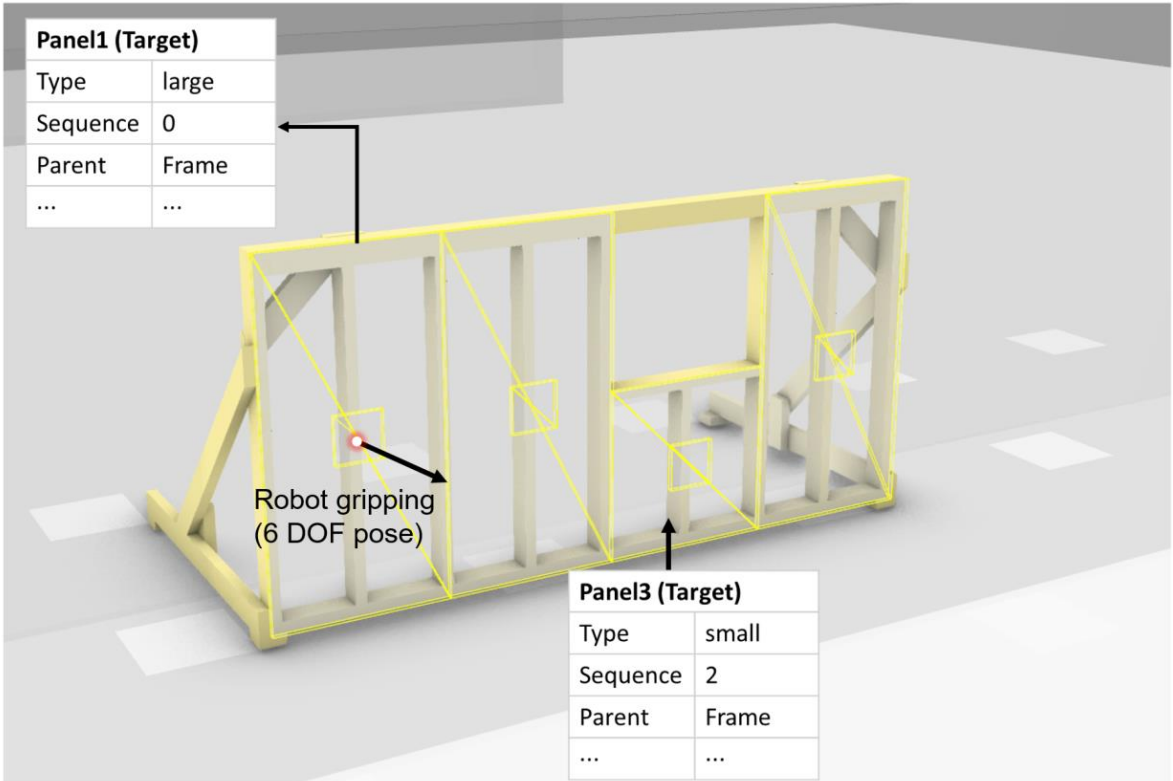

Figure 12: Attribute Information in the BIM to Drive Robotic Construction

### 4.3 Construction Process

To prepare for the construction work, the drywall panels are placed into two stacks in the robot workspace (Figure 9). There is a marker on each panel and the wall frame. Offsets from the markers to the objects' pose indicator points are recorded and input into the system. The marker is not only used for pose estimation but also provides configurable attribute information (e.g., corresponding object type and quantity).

As construction starts, the robot follows a predefined trajectory to scan the environment. After scanning, the robot replaces the as-designed wall frame with the as-built one in the MoveIt planning scene. In the meantime, material poses are inferred from the detected stack location, quantity, and type data attached to the marker. The materials are also added to the MoveIt planning scene as collision objects. When the robot plans motion with materials in hand, the collision object of the corresponding material is attached to the robot end-effector to ensure that the material held by the robot does not collide with the environment or the robot itself. The as-built wall frame and materials are also sent to Unity to be generated in GUI to support user visualization and decision making.

The robot first highlights the next drywall target in the construction sequence in the GUI and asks for the human co-worker's confirmation or adjustment (Figure 13a). After the human co-worker confirms the target, the robot uses the as-built and as-designed deviation of the wall frame to calculate a suggested drywall installation pose. Visualization of the suggested pose is then generated in GUI for the human co-worker's approval or adjustment (Figure 13b). Either in VR or in 3D mode, human co-workers can adjust the camera view to inspect the environment from the perspectives they prefer, while VR offers a more natural and intuitive sight of view control. A video demonstrating the adjustment process can be found at: https://youtu.be/Qdz6fkFpq8s.

After the installation pose is approved, the robot generates the motion plan to first pick up a corresponding type of panel (i.e., large or small) from the detected panel stacks and then places it with the approved installation pose onto the wall frame. Upon request from the human co-worker, a virtual "planning" robot manifests in the GUI and demonstrates the robot motion plan and how the panel is manipulated during the installation process for evaluation (Figure 13c). If the motion plan is approved, the physical robot executes the plan, and the human co-worker can supervise the robot execution process with the synchronized robot and understand the robot status from the messages in the GUI (Figure 13d). After the robot releases the panel, the panel is attached to the wall frame with magnets, and the virtual panel in Unity is changed to the "As-built" layer.

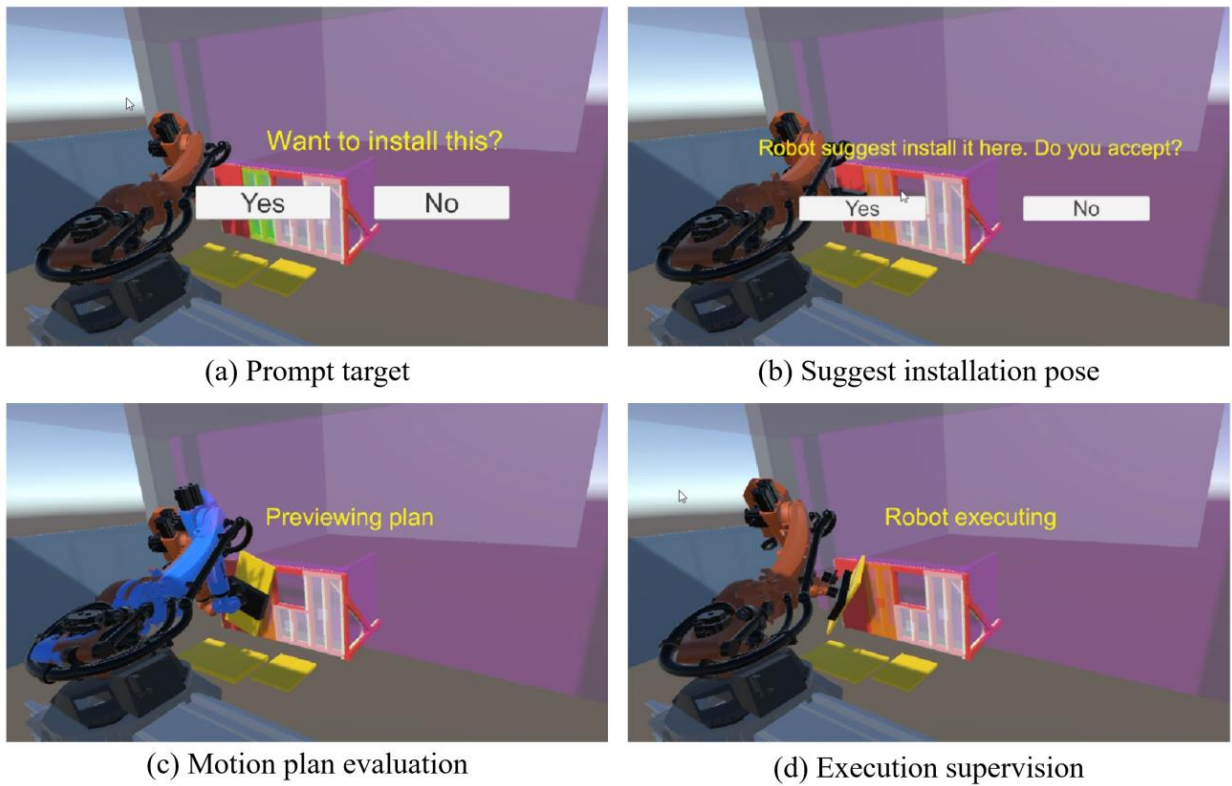

Figure 13: Screenshots of the HRCC Process

The actual installation pose is recorded by the robot and the robot prompts the next panel in the sequence for installation. The quantity of materials in the corresponding panel stack is reduced by one, and the position for the robot to reach the next piece of panel in that stack is updated accordingly. These procedures are repeated until all four pieces of drywall panels are installed. The snapshots of the physical robot drywall installation process during the laboratory experiment are shown in Figure 14. A video demonstrating the process can be found at: https://youtu.be/e4V-hEE-6Tw. After referring to the ISO 12018-1:2011 [87] and the robot manual, the reviewer found 3% of the robot full speed, which set the maximum Tool Center Point (TCP) speed at 60 mm/s (below the specified threshold of 250 mm/s of robot part replacement and reduced speed control), as a comfortable operation speed for the research team to manage and respond to emergencies, minimizing the risk of accidents and preventing potential damage to the laboratory and the robot.

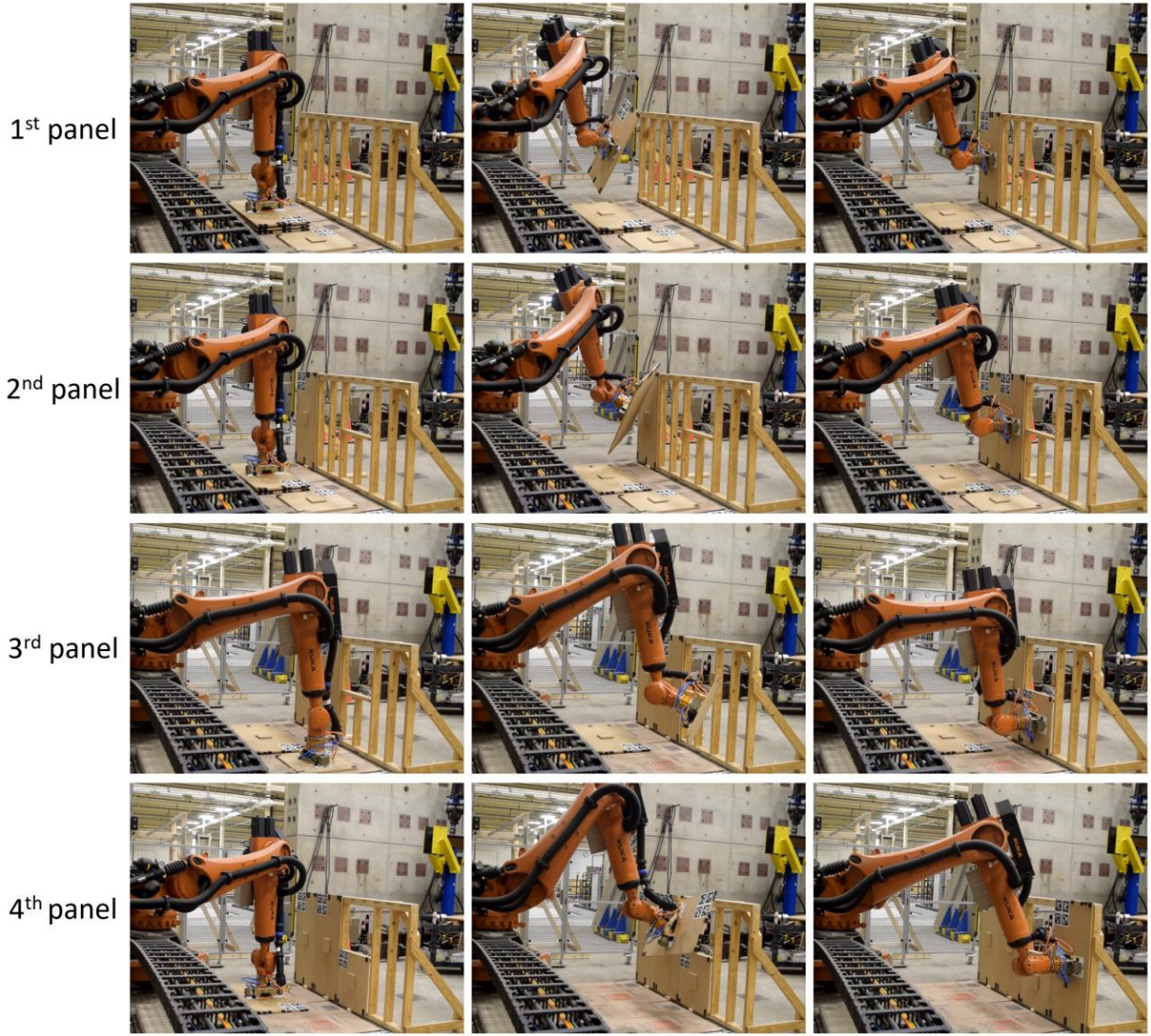

Figure 14: Snapshots of Physical Robot Drywall Installation Process

After all the panels are installed, the as-built condition of the wall frame is sent from ROS to Rhino through Rosbridge using COMPAS as the workspace sensing data [83]. The recorded installation poses of all panels are inferred from the robot end-effector pose and are also sent to be saved in the BIM as the robotic construction data. Both the workspace sensing data and the robotic construction data are saved onto the "As-Built" layer. Lastly, the up-to-date conditions of the panel stacks are sent to Rhino and saved onto the "Material" layer to reflect the quantity and location of the remaining panels on-site. The updated scene in the BIM is shown in Figure 15.

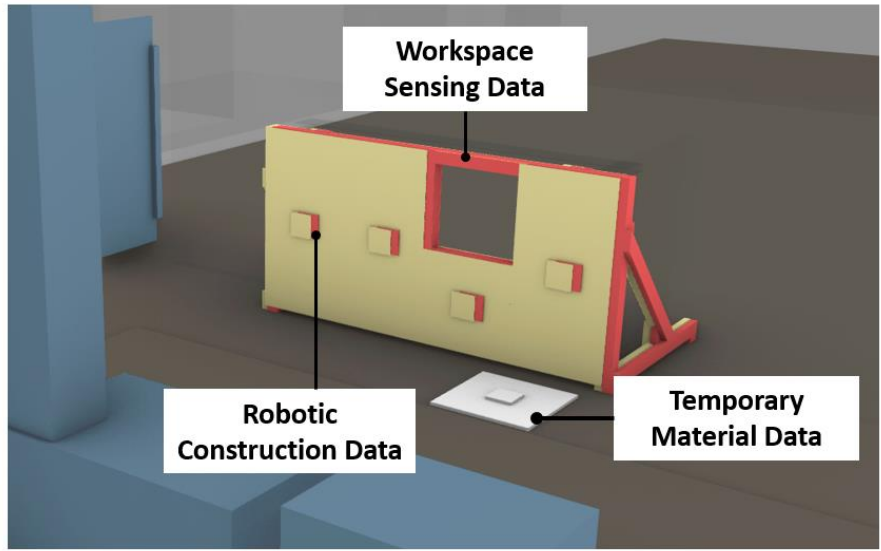

Figure 15: The Updated BIM

# 5 SYSTEM VALIDATION EXPERIMENTS

## 5.1 Physical Experiments for Overall System Evaluation

In Section 4, a drywall installation case study on the physical robot is used to verify the proposed workflow. It is a representative activity in construction to demonstrate the workflow. However, the specific nature of the task does not allow for the exploration of error margins by leaving various spacing through repeated experiments to evaluate system performance. As a result, block pick-and-place experiments that involve a line of four blocks are conducted to validate the system's performance. The BIM repository of the system and the physical experiment setting are shown in Figure 16. Four wood blocks are stacked on the ground floor. A stud is used to represent a nearby object, possibly installed in earlier construction states. AprilTag markers are positioned near the block stack and the stud for component identification and localization. The blocks are expected to be placed in a line adjacent to the stud. Block pick-and-place is selected as the task here because it is fundamental yet representative of construction activities. On one hand, its simplicity facilitates repetitive testing and control over experimental conditions, such as the spacing between blocks and end stud positioning. On the other hand, it requires precision and adaptability to deviations typically needed for construction tasks thereby allowing us to evaluate system performance in a representative construction context.

The robot first scans the environment to localize the block stack and stud through the ApirlTag markers placed near these components. Then, it needs to first pick up a wood block from the stack of blocks and place it alongside the stud placed on the ground. If the robot finds the stud takes up the space for the planned block placement target, it will suggest offsetting the block placement target to avoid collisions. If the stud does not occupy the space of the blocks, the robot will follow the original plan and will not automatically make adjustments, unless instructed by the human supervisor.

In order to increase system tolerance to errors and prevent damage to experiment materials and the robot, gaps of different sizes (10 mm, 5 mm, 3 mm, and 1 mm) are left between blocks.

The gap sizes are initiated at 10 mm and are gradually reduced to approximately half of the previous value to test the margin of system precision. The gripper releases and drops the block 2 mm above the ground floor, leaving room for possible vertical errors. For each gap size, 10 trials of picking and placing all four blocks were carried out. The wood stud was deliberately placed in collision with the planned target for 5 out of the 10 trials to test the robot's capability to resolve deviation. Since the task has lower risks compared to drywall installation, the robot is operating at a slightly higher speed of 7% of the robot's full speed (140 mm/s) as the research team's comfortable speed for experiment management and emergency handling.

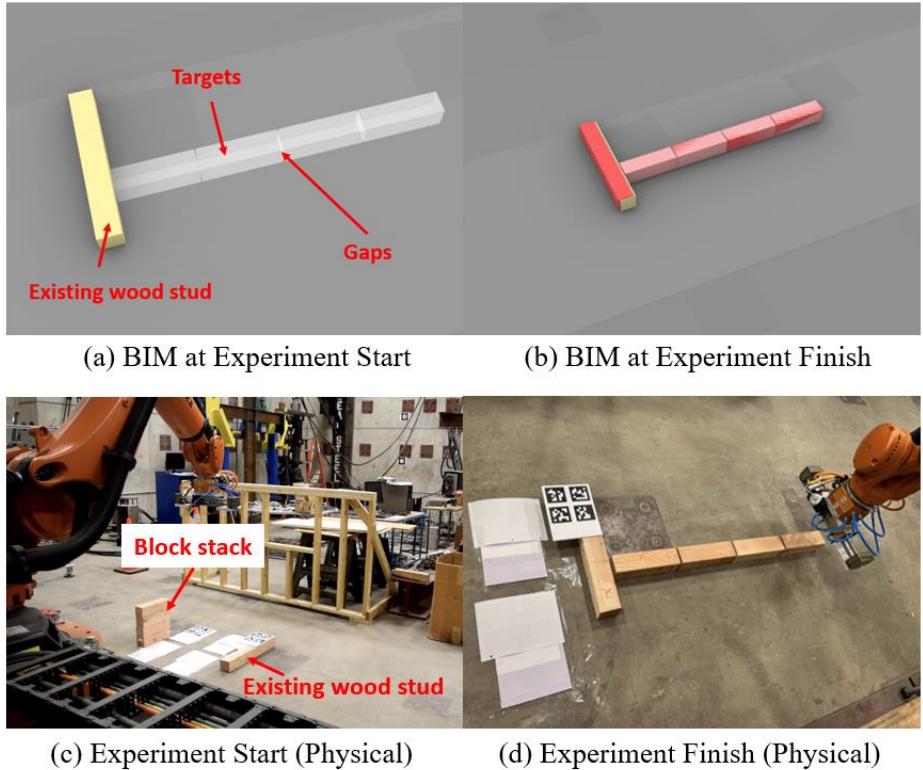

Figure 16: Experimental Settings

During the experiment, the human co-worker interacts with the robot through the 3D interface. The goal of this experiment is to verify the overall performance of the proposed system when the human workers completely rely on robot suggestions. Therefore, the human co-worker agrees with all robot suggestions and does not perform any adjustment or manual intervention during the construction process. However, when the co-worker feels the robot planned manipulation trajectory is not optimal (e.g., taking extra rotations), the co-worker would request the robot to generate a new motion plan for evaluation. The number of replanning requests from the human co-worker is recorded. The trial is counted as a failure if the robot does not place all four blocks successfully.

The success rate and the number of replanning requests for different gap sizes are shown in Table 3. "Successful placements" means the number of blocks successfully placed by the robot without any collision. Once a block fails, the trial ends and the rest of the blocks in the four-block line are not placed, resulting in more than one block not being successfully placed with one failure case. For example, failure of the first block placement will end the trial and cause all four blocks not being successfully placed. On average, a successful pick-and-placement trial of four blocks takes 217.31 seconds. 56.38 seconds are used for human co-worker's decision-making, such as confirming the target and previewing the motion plan. The average time taken by robot computation and execution is 160.93 seconds.

Table 3 Block Pick-and-Place Experiment Results

| Size of gap | Success rate (%) | Replan requests / Successful placements | Reason for failure (occurrence) |
| --- | --- | --- | --- |
| 10 mm | 100 | 7 / 40 | |
| 5 mm | 90 | 2 / 38 | Hit ground (1) |
| 3 mm | 90 | 6 / 36 | Collide with stud (1) |
| 1 mm | 60 | 5 / 26 | Collide with stud (3) Hit ground (1) |

## 5.2 Simulation Experiments for Validation of Nearby Object Deviation Adaptation

In order to specifically assess the system's capability to autonomously adapt to deviations of nearby objects and to accurately trace source errors in physical experiments, a block pick-and-place experiment is conducted in Gazebo simulation, mirroring the settings of the experiments in Section 5.1. While physical experiments evaluate the performance of the overall system, simulation enables precise tracking of component positions and orientations for performance assessment and allows isolating other error sources to focus on the evaluation of specific system components. It should be noted that the Gazebo simulation has been demonstrated to be replicable on the physical robot system used in this study with high accuracy [88]. In this experiment, the robot is provided with ground truth poses of tags to eliminate errors arising from component localization. The stud is intentionally offset towards the target block placement location, necessitating robot adaptation to avoid collisions. Across 10 trials, the deviation distance of the stud is set to a randomly generated value between 0 and 0.2m. No gaps are left between blocks or between a block and the ground. As the focus is to evaluate the robot's capability for autonomous deviation adaptation, all the robot's suggested solutions are accepted without human intervention.

Ground truth poses and actual locations of each of the four blocks in the simulation environment are recorded and compared. To avoid errors due to movements of the loose objects post-handling, actual block locations are recorded before the robot releases them. Figure 17 shows examples of the experiment results. The stud is in white and the blocks are in wood material in Gazebo simulation. Deviations of the stud and the adapted installation pose can be visualized in the digital twin interface. Errors of all blocks across 10 trials are analyzed in Table 4. Both the Root Mean Square Error (RMSE) and the Standard Deviation (STD) are close to zero, demonstrating the effectiveness of the proposed deviation adaptation method for the nearby object deviation situations.

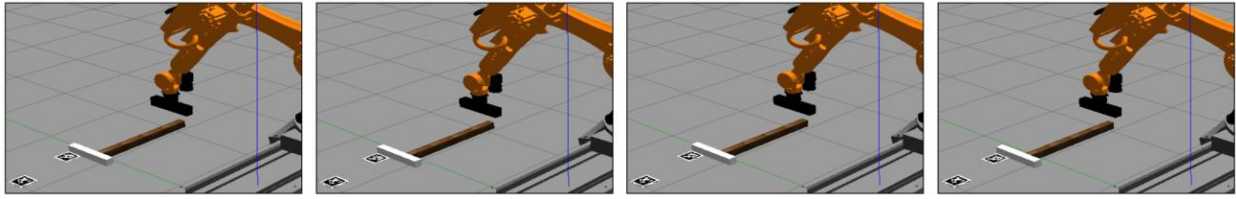
(a) Gazebo simulation

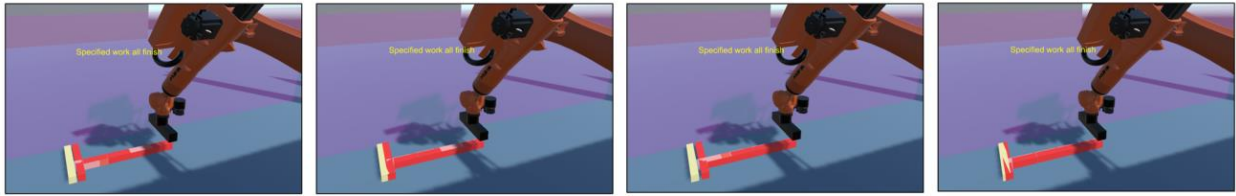
(b) Digital twin interface (yellow: as-designed; red: as-built)

Figure 17: Examples of Block Pick-and-Place Simulation Experiments

Table 4 Error Analysis for Block Pick-and-Place Simulation Experiment

|  | Position errors | | | Orientation errors | | |
| --- | --- | --- | --- | --- | --- | --- |
|  | x | y | z | roll | pitch | yaw |
| RMSE | 1.657e-04 | 5.163e-05 | 4.864e-04 | 2.033e-04 | 1.289e-04 | 2.426e-04 |
| STD | 1.651e-04 | 5.023e-05 | 4.534e-04 | 1.939e-04 | 1.266e-04 | 2.423e-04 |

### 5.3 Simulation Experiments for Validation of Parent Deviation

To assess the system's capability to autonomously adapt to the parent deviations, a drywall installation experiment with settings replicating the case study in Section 4 is conducted in Gazebo simulation. Ground truths of tag poses are provided to the robot to eliminate the errors arising from component localization. The frame is intentionally offset in both position and orientation to necessitate the adjustment for subsequent installation of panels. Across 10 trials, the deviation distances of the frame on the X and Y axis are set to randomly generated values between 0 and 0.2m. The yaw orientation deviation of the frame is randomly generated between 0 and 15 degrees. This range ensures the frame stays within the robot workspace and the panel installation pose are valid and reachable by the robot. The robot's suggested solutions are always approved without human intervention.

The experiment compares the actual poses of each of the four panels in the simulation environment with their ground truth poses. Figure 18 shows examples from the experiment trials. The deviations of the frame, along with the original and adapted installation pose, can be visualized in the digital twin interface. The errors of all panels across 10 trials are analyzed in Table 5. Similar to the nearby object deviation situation, both the RMSE and the STD are close to zero, demonstrating the effectiveness of the proposed deviation adaptation method for the parent deviation scenarios.

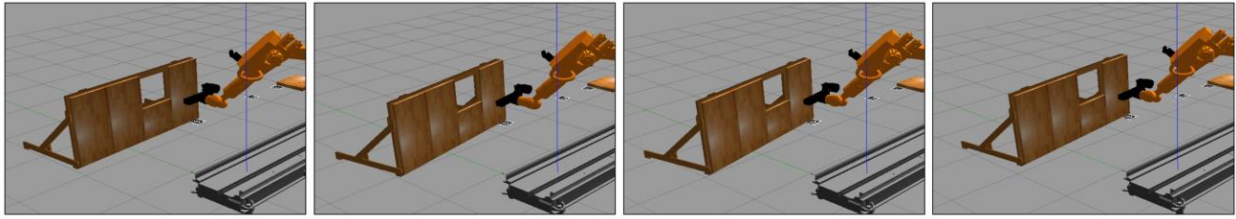

(a) Gazebo simulation

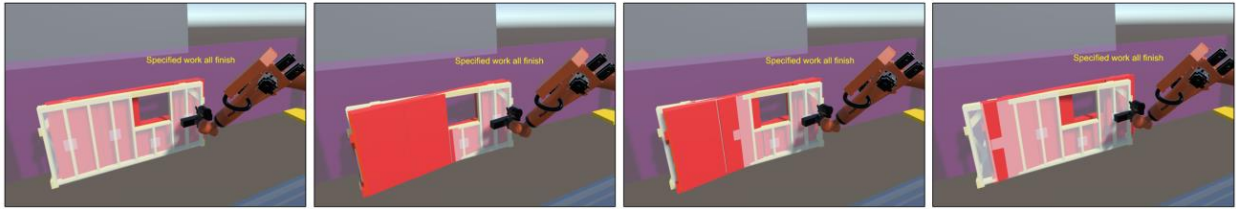

(b) Digital twin interface (yellow: as-designed; semi-transparent: original targets; red: as-built)

Figure 18: Examples of Drywall Installation Simulation Experiments

Table 5 Error Analysis for Drywall Installation Simulation Experiment

|      | Position errors | | | Orientation errors | | |
| --- | --- | --- | --- | --- | --- | --- |
|      | x | y | z | roll | pitch | yaw |
| RMSE | 8.347e-04 | 8.513e-05 | 1.160e-04 | 6.046e-05 | 3.759e-05 | 1.376e-05 |
| STD  | 8.211e-04 | 8.311e-05 | 6.481e-05 | 3.340e-05 | 2.268e-05 | 1.288e-05 |

## 6 DISCUSSION

In order to successfully perform the construction work with the proposed system, the robot needs to adequately localize components in the construction environment, make decisions and suggestions to adapt to uncertainties, and accurately reach specific positions and manipulate components. Errors in any of these aspects will disrupt the workflow. The low errors from experiments in simulation indicate that the robot can accurately make adaptations and find the appropriate pose to place the target. The block pick-and-place experiment with a physical robot is conducted to evaluate the overall system performance. It is observed that once the first block is placed, the rest of the blocks are placed without collision with each other. It indicates that after the robot accurately determines the adapted installation poses, it executes the plan with high precision.

Most failure cases are caused by the first block colliding with the stud while being placed (Figure 19). There are also two cases of failure where the blocks were moved too close to the ground and the robot sensed excessive force on its end-effector. These failures are caused by errors in component localization. AprilTag markers are used for component localization, which is a low-cost and easy-to-deploy solution that provides relatively high localization accuracy and is robust to various environmental conditions [89,90]. However, it still introduces certain localization errors, which can lead to collisions when the tolerance is very low. For example, in the drywall installation case study, the panel cannot be firmly held onto the frame if a minor gap between magnets exists. Since the marker is installed on one end of the frame, orientation errors in marker detection are amplified by the long distance and cause a larger position offset when installing the panel on the other end of the frame. This results in task failure in some cases during the implementation process. Fortunately, with the evolving object detection algorithms and hardware (e.g., LiDAR) [91,92], the accuracy of component localization is increasing to meet the needs of precise construction tasks.

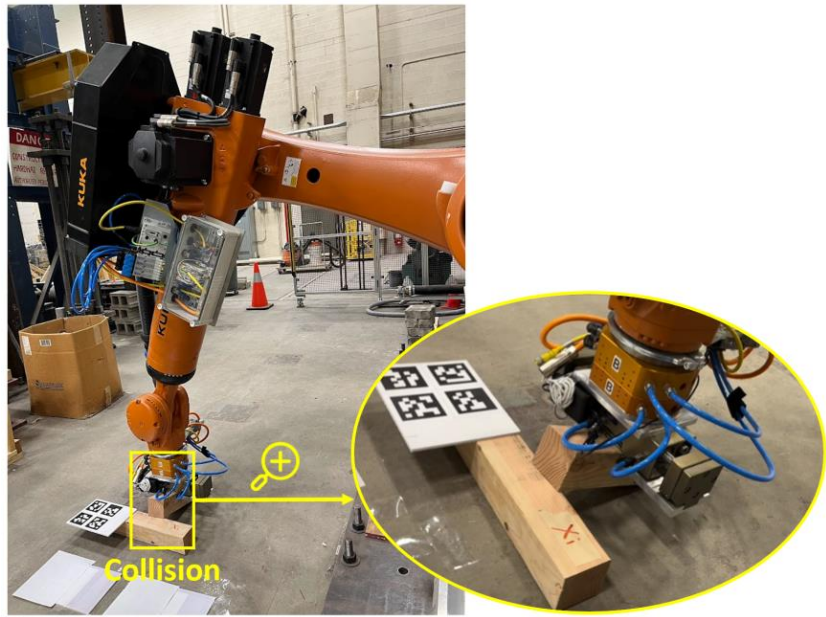

Figure 19: Collision with Stud during Block Placement

Considering the visual localization error, the success rate of the block pick-and-place experiment is found to be particularly high. Several reasons may lead to this result. First, all the blocks are stacked at one location, and thus the same offsets are maintained for all blocks. Therefore, collisions between blocks are avoided provided that the digital twin system successfully adapts to the nearby object deviation. Second, the gripper compensates for certain localization errors when closing the jaws to grab the block at its horizontal center. Third, only the localization error in a certain direction can cause task failure. For example, the block will only collide with the stud if the stud is in the way of the block target pose but the robot thinks the stud is still far away and does not make any adjustments to avoid the collision. Fourth, the robot drops the blocks onto the ground at a height of 2 mm, which increases the vertical tolerance of the task. Even if the robot drops the block from a slightly higher or lower position, the block placement is still considered successful.

This study leverages the meshes as BIM component geometry and employs the proposed layer structure to generate the interactive digital twin through template programs in Unity, demonstrating that the digital twin can be generated provided the components are placed on the designated layers. Different from existing studies that parse component IFC files for robot motion planning, this study directly loads the meshes of BIM components as collision objects in the MoveIt planning scene and uses the robot gripping attribute data from the BIM to generate collision-free motion plans. Therefore, the manipulation targets or other objects are not required to conform any specific shape or follow a particular IFC structure to be processed by the system. The BIM update process leverages the existing component geometries in BIM and generates as-built data by offsetting the existing components based on the subscribed pose information. This approach avoids the challenges associated with creating diverse and complex geometries. These features enhance the framework's flexibility for various types of construction assembly tasks. Two distinct tasks, drywall installation and brick pick-and-place, are implemented in this study for verification. Since a variety of other construction tasks can be similarly composed with the underlying elemental motions and large-size object manipulation methods from these example tasks, it can thus be readily adapted to the developed methods directly or with minor modifications [37,85]. However, it should be noted that the proposed system is primarily designed for rigid objects and is less effective for tasks involving fluid or deformable materials such as freshly mixed concrete or waterproofing membranes.

During the physical experiment, several limitations are observed, and future research directions are identified. First, as discussed above, although the robot can localize objects with up to millimeter-level accuracy, it is not sufficient for some construction tasks that require high precision. It also takes time to set up and for the robot to scan fiducial markers on all related components in the workspace. A more direct and precise approach for the robot to perceive the environment, continually track components, and subsequently update BIM based on the detected installation pose of components should be considered in future developments.

Second, because of the laboratory condition, only one robotic arm is used, and the panels are grabbed by a 2-jaw gripper in the drywall installation case study. One more robotic arm can be included to fasten panels onto the wall frame, and a vacuum gripper can be used to grab panels so that cubic handles are not needed.

Third, information to support robotic construction, such as robot gripping pose and construction sequence, is manually created in the BIM. Future work should integrate computational design into the system, which can automatically generate detailed digital fabrication information, construction sequence, and gripping plans for components [93].

Fourth, some motion plans generated by the robot are valid but not desirable, and replanning is preferred. Sometimes, the robot manipulates material extremely close to other objects on-site. While no collision occurs, these situations cause high mental stress on human co-workers when the workpiece is large. Future studies may consider applying force feedback control and reinforcement learning for more trustworthy and desirable component manipulation [94,95].

Lastly, for experimental purposes, the robot's working speed is deliberately set to a slow pace (3% or 7% of its full capacity) to minimize risks such as object falls, collisions, and robot damage, and to emphasize the safety of the research team conducting the experiments. This results in extended work times. In the future, the efficiency of the workflow can be improved by applying faster robot speed, foregoing the preview step for low-risk operations, and eliminating the need for human approval when the confidence level of robot decision is high. Future studies will benefit from a comprehensive efficiency assessment that analyzes various impact factors.

## 7   CONCLUSIONS

This study proposes a BIM-driven HRCC workflow that addresses technical solutions ranging from the preparation stage to the end of the construction work, enabled by a closed-loop digital twin framework. The proposed framework offers several significant improvements to the previous I2PL-DT as well as other independent contributions. First, it presents a BIM framework that supports HRCC. The BIM contains the attribute and geometric information human workers and robots need for construction, and the predefined layer structure provides a unified standard to interface different BIM projects with the interactive digital twin, thereby improving BIM interoperability. Second, an automatic approach for generating interactive digital twins for HRCC is proposed, which is enabled by a template-based Unity program and the predefined layer structure in the BIM. Third, this study introduces an approach for component placement deviation adaptation using a combination of as-designed data from the BIM and perceived as-built information. Lastly, the construction site as-built information is sent to the BIM to be recorded, forming a closed-loop system. By closing the loop, the BIM is updated with as-built data to support decision-making and automation in subsequent construction, operation, and maintenance of a facility.

Physical and simulation experiments are conducted to identify the effort needed to enable a physical construction robotic system, verify and validate system performance, and recognize limitations for future improvements. Overall, through the integration of the BIM, the proposed system not only improves construction work quality but also increases the robot's capability in lower-level task planning thereby reducing human workers' planning efforts.

## ACKNOWLEDGMENTS

The authors would like to acknowledge the financial support for this research received from the U.S. National Science Foundation (NSF FW-HTF-P #2025805 and FW-HTF-R #2128623). Any opinions and findings in this paper are those of the authors and do not necessarily represent those of the NSF.